\documentclass[pdflatex,sn-nature]{sn-jnl}

\usepackage{graphicx}
\usepackage{multirow}
\usepackage{array}
\usepackage{amsmath,amssymb,amsfonts}
\usepackage{booktabs}
\usepackage{url}
\usepackage{float}
\usepackage{pdfpages}

\usepackage{etoolbox}
\makeatletter
\patchcmd{\@maketitle}{\removelastskip\vskip20pt\nointerlineskip}{\removelastskip\vskip12pt\nointerlineskip}{}{\typeout{PATCH1 FAILED}}
\patchcmd{\@maketitle}{\vskip20pt\else}{\vskip20pt\else}{}{}
\patchcmd{\@maketitle}{\global\punctcount\aucount\vskip20pt}{\global\punctcount\aucount\vskip12pt}{}{\typeout{PATCH2 FAILED}}
\patchcmd{\@maketitle}{\removelastskip\vskip24pt}{\removelastskip\vskip14pt}{}{\typeout{PATCH3 FAILED}}
\patchcmd{\@maketitle}{\removelastskip\vskip24pt}{\removelastskip\vskip14pt}{}{\typeout{PATCH4 FAILED}}
\makeatother

\begin{document}

\title[AtmosCoder-Bench]{Execution-grounded evaluation reveals hidden failures in language-model calculations for environmental science}

\author[1]{\fnm{Maohao} \sur{Ran}}\equalcont{These authors contributed equally to this work.}
\author[1]{\fnm{Chendong} \sur{Ma}}\equalcont{These authors contributed equally to this work.}

\author[2]{\fnm{Yanting} \sur{Zhang}}
\author[1]{\fnm{Dailing} \sur{Jiang}}
\author[3]{\fnm{Yusen} \sur{Huang}}
\author[1]{\fnm{Meng} \sur{Gao}}
\author*[1]{\fnm{Jun} \sur{Song}}\email{junsong@hkbu.edu.hk}

\affil[1]{\orgdiv{Department of Geography},
  \orgname{Hong Kong Baptist University},
  \orgaddress{\city{Hong Kong}, \country{China}}}

\affil[2]{\orgname{Hong Kong University of Science and Technology},
  \orgaddress{\city{Hong Kong}, \country{China}}}

\affil[3]{\orgname{The Chinese University of Hong Kong (Shenzhen)},
  \orgaddress{\city{Shenzhen}, \country{China}}}

\abstract{Large language models (LLMs) are increasingly proposed for quantitative work in the environmental sciences, yet their computational abilities in this domain remain underexplored. Existing evaluations typically score only final answers, leaving the calculation process unobserved. Here we introduce AtmosCoder-Bench, an execution-grounded benchmark for atmospheric science in which models must produce answers through executed code, making the computational process visible and auditable. The benchmark is constructed through a transferable, semi-automated pipeline (436 problems, 3,910 variants, 7,029 graded quantities), with each problem validated to be unambiguous and uniquely verifiable. We find that (i) multiple-choice formats inflate measured accuracy by 12 to 39 percentage points; (ii) many failures arise not from missing knowledge, but from models failing to apply known formulas and constraints consistently throughout multi-step computation; and (iii) even frontier models remain weak when task-specific conditions invalidate familiar methods, often reverting to canonical solution patterns instead of adapting to the relevant physical regime. Execution grounding also reduces token use and yields robust performance under numerical and paraphrase perturbations. Together, these results show that, for frontier models, the remaining bottleneck lies less in raw calculation than in domain-specific judgement and the reliable application of known principles.}

\keywords{large language models, scientific computation, benchmarking, atmospheric science, code generation\\[4pt]
\textbf{GitHub:} \url{https://github.com/acodercat/AtmosCoder-Bench}}

\maketitle

\section{Introduction}\label{sec:intro}

Atmospheric science and the broader environmental sciences are among the most societally consequential and computationally demanding scientific domains. The systems they describe are governed by coupled, nonlinear processes, observed through vast and heterogeneous data streams, and studied under exacting accuracy requirements \cite{bauer2021digitaltwin,bauer2024digitaltwins,rolnick2022tackling,reichstein2019deep,irrgang2021towards,lam2023learning,bi2023accurate}, since their conclusions inform climate projection, air-quality management, and disaster response, where small quantitative errors carry large real-world costs \cite{ipcc2021ar6wg1,cohen2017estimates,hallegatte2012costeffective,ugarov2023lives}. Large language models (LLMs) are now applied across this domain \cite{vaghefi2023chatclimate,lin2023geogalactica,hadid2024geoscience,bi2024oceangpt,song2025airgpt,pantiukhin2025earthscience}, and increasingly not merely to retrieve or summarise knowledge but to perform its quantitative core: setting up and analyzing numerical problems \cite{cllmate2024,varambally2025zephyrus,wang2026climagent}, designing computational experiments \cite{thulke2024climategpt,schimanski2023climatebertnetzero,kao2025univearth}, and carrying out the multi-step calculations on which domain reasoning depends \cite{wang2024scibench,ma2024sciagent,gao2025airqualitycasestudy,xu2026envllm}. Yet this same promise raises a question that is now central to the field and remains largely unresolved: when an LLM is asked to perform complex computations as a research assistant in a professional environmental-science setting, can its answers be trusted? Answering this question demands domain benchmarks that measure whether such models are genuinely accurate when a task requires scientific computation, and it therefore places unusually high stakes on how those benchmarks are designed.

A growing body of benchmarks now evaluates LLMs across this domain, spanning climate-knowledge question answering (QA) \cite{spokoyny2023climabench,manivannan2025climaqa,bulian2023assessing,mutalik2025cpiqa,kurfali2025climateeval,li2025atmossci}, policy analysis and management assistance \cite{dai2025oneatmos,manivannan2026generative,mullappilly2023arabic,wang2026climagent}, related environmental analysis \cite{guo2024elle,huang2024enviroexam,he2025esgenius,jegham2025hungry}, and the wider Earth-science field \cite{bi2024oceangpt,xu2025earthse,deng2024k2,kao2026towards}. Benchmarks that systematically evaluate numerical-analysis ability, however, remain limited. There are two representative works: AtmosSci-Bench, which uses a symbolic engine to expand graduate-level textbook problems into large families of multiple-choice questions (MCQs) across five subfields \cite{li2025atmossci}, and OneAtmos-Bench, which provides 994 expert-validated QA pairs spanning true/false, calculation, and open-ended formats \cite{dai2025oneatmos}. Yet both of them are constrained: the MCQ format has been shown to be unreliable \cite{zheng2024notrobust,balepur2024artifacts,xue2024symbolbinding,gupta2024answerorder,chizhov2025hellaswag}, while OneAtmos-Bench's reliance on human experts to compute and validate answers inherently caps the number of computational items it can contain. Both evaluators assess only the final answer, making it impossible to separate the measurement of reasoning ability from answer generation. LLMs are known to make substantive errors in calculation-intensive tasks, such as arithmetic slips, flawed deduction, misunderstanding of niche concepts, and factual mistakes \cite{mirzadeh2024gsmsymbolic,wang2024scibench,he2024olympiadbench,feng2025physics,wang2026frontierscience,ye2025mmscibench}, and recent error analyses attribute most failures to reasoning rather than knowledge \cite{ye2025mmscibench,wang2026frontierscience}. Assessing how these models perform on the domain's own quantitative tasks is therefore important.

More fundamentally, prior efforts share three practical weaknesses: computation is \emph{opaque}, since only the final answer is scored; ground truth is \emph{human-intensive}; and open-ended responses are graded by \emph{LLM judges} \cite{dai2025oneatmos,ye2025mmscibench,wang2026frontierscience} or symbol matching \cite{chow2025physbench}, a non-deterministic, non-reproducible grading surface. During our research, we identified a further, more insidious failure mode: models can misrepresent their execution process silently. They may perform one calculation while claiming another, or even claim to have computed a result while fabricating intermediate steps without actually executing the calculation. Because no explicit error is raised, answer-only grading cannot distinguish such cases from ordinary calculation failures. This risk becomes more important as domain researchers without programming expertise increasingly rely on LLM-based calculations. Both representative benchmarks contain traces of this issue. In AtmosSci-Bench, our re-examination found that some recorded solutions describe code execution that does not appear to have taken place; they were graded as ordinary calculation failures (Appendix~K of that work) \cite{li2025atmossci}. It is an understandable outcome, since answer-only grading offers no signal by which such cases could be distinguished. In OneAtmos-Bench (Supplementary Table S6 of that work), GPT-4o, when asked to establish the order of accuracy of a finite-difference advection scheme, claimed to have performed a Taylor-series expansion and collected the resulting error terms, yet actually arrived at the answer through heuristic judgement rather than the claimed computation \cite{dai2025oneatmos}. These cases, therefore, became the major motivation for this work: we set out to build a benchmark framework in which models are compelled to perform code execution, with a process that can be seen and audited, thereby preventing such hallucinations, or at least, making them visible. Program-aided prompting has previously used code execution to improve LLM accuracy on numerical reasoning tasks \cite{gao2023pal,chen2023pot}; here execution serves instead as the measurement and audit surface, paired with certified numeric ground truth. This enables a more accurate evaluation of models' actual computational ability.

We built \textbf{AtmosCoder-Bench}, an execution-grounded benchmark for atmospheric computation that pairs verifiable ground truth with a visible, auditable calculation process, using a pipeline extensible to the broader field of environmental analysis. Problems are harvested through an automated extraction pipeline, and every reference answer is frozen through both frontier-model validation and human expert verification to establish the ground truth. Each model is then required to write and execute code to produce the answer, making the computational process auditable; outputs are graded against a single deterministic numerical solution. This exact and transparent evaluation allows us not only to measure the actual computational abilities of LLMs, but also to identify where they fail during calculation, and quantify limitations. To our knowledge, this is the first execution-grounded benchmark with certified numeric ground truth in the environmental domain. For the benchmark database, we extracted 436 quantitative QA items from 13 undergraduate- and graduate-level textbooks, organised into 10 representative categories of atmospheric science and three difficulty levels. We generated five numeric variants for every perturbable problem and five paraphrase variants for every problem, evaluated each model configuration over three independent runs, and re-graded answers at several tolerance levels to ensure a more robust and fair assessment.

Our principal findings are as follows:
\begin{itemize}
  \item \textbf{We quantify the extent to which multiple-choice formats inflate measured ability.} On identical problems, option-mode accuracy exceeds computed open-answer accuracy by 20--39 points, with a 12--27-point gap remaining after defective answer keys are removed.
  \item \textbf{LLMs exhibit a persistent gap between knowing the right rule and applying it correctly.} In many cases, models can identify the correct formulas or rules but fail to apply them consistently, with recurring failures such as truncating long derivations, losing track of constraints across multiple steps, or claiming to use a right concept or parameter while actually using another. Stronger reasoning models reduce, but do not eliminate these failures.
  \item \textbf{The remaining ceiling of frontier models lies in domain-specific judgement.} Even strong models often fail to recognize when task-specific conditions invalidate a familiar method. They tend to fall back on a canonical solution pattern while disregarding the specific physical or geographical conditions of the problem, even when explicitly instructed otherwise. What is systematically missing is the professional judgement to adapt a method to the unique conditions of the task at hand, a capacity for which domain expertise remains irreplaceable.
\end{itemize}

We further find that most models remain consistent across variants in execution-based pipelines. Execution grounding can reduce truncation from natural-language reasoning and better isolates computational ability, while prose-only evaluation spends about 1.5 times as many tokens over a full run and may fail to return an answer. Execution fabrications exist but are rare, observed only in smaller models, yet their silent nature still warrants dedicated detection and monitoring. We also applied the identical construction pipeline to four additional environmental domains (hydrology, environmental chemistry, ecology and biogeochemistry, and soil mechanics); the resulting 131 problems separate the five evaluated models in exactly their atmospheric-benchmark order (Spearman $\rho=1.00$), indicating that it is the construction method, rather than the subject matter, that produces a discriminative measurement. Finally, we find that prompt formulation affects performance, but the effect is concentrated in weaker models and operates primarily through executability rather than physical reasoning. Execution grounding disentangles these factors and makes failures directly traceable, rendering the calculation process fully auditable, which is a key advantage of this architecture.

\section{Results}\label{sec:results}

\subsection{An execution-grounded benchmark}

\begin{table}[t]
\centering
\small
\caption{\textbf{Composition of AtmosCoder-Bench by category.} 436 base problems from 13
established textbooks spanning introductory to advanced level (undergraduate to graduate), each labelled with one of ten
atmospheric--environmental categories. \emph{Follow-up queries} counts the additional sub-questions a
problem poses beyond its first (a problem asking for parts (1)--(3) contributes two); every
sub-question is graded independently. Each perturbable problem carries five parameter-perturbed
\emph{numeric variants} (346 problems; contamination resistance) and every problem five
\emph{paraphrase variants} (phrasing robustness; answer unchanged).}
\label{tab:stats}
\begin{tabular}{lrrrr}
\hline
Category & Problems & Follow-up & Numeric & Paraphrase \\
         &          & queries   & variants & variants \\
\hline
Atmospheric dynamics       & 116 & 47 & 520 & 580 \\
Atmospheric thermodynamics &  89 & 59 & 395 & 445 \\
Atmospheric chemistry      &  51 & 76 & 160 & 255 \\
Air quality                &  37 &  8 & 130 & 185 \\
Boundary layer             &  32 & 16 & 145 & 160 \\
Cloud physics              &  29 & 20 & \phantom{0}85 & 145 \\
Atmospheric radiation      &  25 & 38 & 105 & 125 \\
Atmospheric aerosols       &  24 & 16 & \phantom{0}75 & 120 \\
Climate dynamics           &  19 &  8 & \phantom{0}55 & \phantom{0}95 \\
Observation \& modeling    &  14 & 10 & \phantom{0}60 & \phantom{0}70 \\
\hline
\textbf{Total} & \textbf{436} & \textbf{298} & \textbf{1{,}730} & \textbf{2{,}180} \\
\hline
\multicolumn{5}{l}{\emph{Difficulty levels:} low 44 \quad medium 258 \quad high 134} \\
\hline
\end{tabular}
\end{table}

\begin{figure}[t]
\centering
\includegraphics[width=0.9\textwidth]{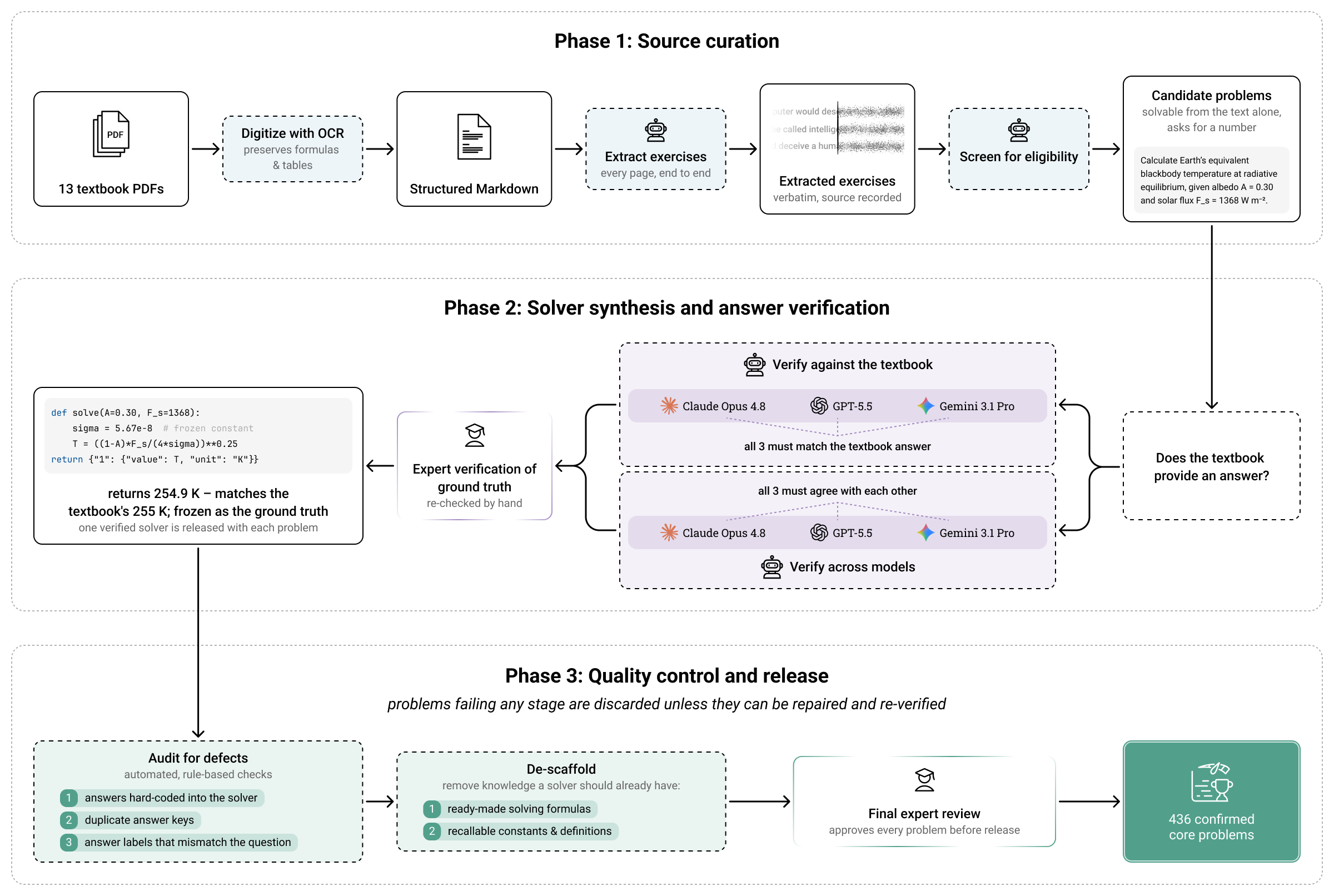}
\caption{\textbf{The AtmosCoder-Bench admission pipeline.} Textbook items pass a sequence of gates---%
extraction for self-contained numeric answerability, an executability gate, cross-vendor
independent-derivation agreement (textbook-anchored or consensus-frozen), an AST$+$perturbation
faithfulness audit, a well-posedness audit and triple-gated de-scaffolding---before certified,
contamination-resistant variants are generated. Items failing a gate are removed (right), so the
admitted set carries verifiable, execution-grounded ground truth.}
\label{fig:pipeline}
\end{figure}

AtmosCoder-Bench is an execution-grounded benchmark: every answer is produced by a verifiable computation rather than asserted, and is graded against deterministic ground truth certified by independent cross-model derivation. The benchmark comprises 436 self-contained computational problems drawn from 13 established textbooks spanning introductory to advanced levels across the atmospheric and environmental sciences (Supplementary Table~S4), grouped into ten physical categories and three difficulty levels: tasks requiring only simple retrieval and statistics are marked low, those requiring analysis and application are marked medium, and those demanding advanced multi-step calculation, modelling, or deduction are marked high (Table~\ref{tab:stats}; Supplementary Figure~S3). Each question contains only the problem text, with no formulas, expected answers, or other solution cues; models must return a standard-library \texttt{solve()} function, so that it is the interpreter, not the model's prose, that produces the graded number, which is scored against the reference within a 5\% relative tolerance.

The database was built with a three-phase semi-automated pipeline (Figure~\ref{fig:pipeline}). Candidate problems were extracted from textbooks, cleaned, and screened for eligibility; ground-truth answers were taken from the textbook where available or otherwise certified by the agreement of three independent frontier models, followed by manual verification; a final quality-control stage removed defective, ambiguous, or physically impossible items and stripped any content that would let models bypass the calculation, such as ready-made solving functions. The resulting task set is clean, professionally solvable, and deterministically single-answer. Full construction details are given in Methods.

To test robustness and distinguish genuine calculation from pattern matching, each question was additionally subjected to two types of variation: numeric variants, in which the numbers are replaced and the ground truth recomputed, and paraphrase variants, in which the question is reworded without changing its meaning or numbers. Since some questions admit no numeric variation, the benchmark contains 1730 numeric variants and 2180 paraphrase variants (Table~\ref{tab:stats}).

\subsection{Restrictive answer formats inflate measured accuracy}

\begin{table}[t]
\centering
\small
\caption{\textbf{Multiple-choice inflation: option mode versus computed answers on identical problems.}
$\Delta$ is chance-corrected option accuracy minus computed accuracy at the most permissive
$\pm20\%$ tolerance; ``(r)'' denotes a reasoning configuration. $^{*}$Option accuracy from the
source benchmark\cite{li2025atmossci}.}
\label{tab:mcq}
\begin{tabular}{lcccccr}
\hline
\multirow{2}{*}{Model} & \multicolumn{2}{c}{Option mode} & \multicolumn{3}{c}{Code mode (grading tolerance)} & \multirow{2}{*}{$\Delta$} \\
 & raw & chance-corr. & $\pm5\%$ & $\pm10\%$ & $\pm20\%$ & \\
\hline
Gemini-3.1-Pro (r)    & 94.5\%       & 92.7\% & 60.0\% & 61.0\% & 61.9\% & $+30.8$ \\
DeepSeek-R1 (r)       & 88.5\%$^{*}$ & 84.7\% & 53.0\% & 54.6\% & 55.7\% & $+29.0$ \\
GPT-5.5               & 80.1\%       & 73.5\% & 49.6\% & 50.6\% & 53.0\% & $+20.5$ \\
DeepSeek-V4-flash     & 88.1\%       & 84.1\% & 43.3\% & 44.0\% & 45.8\% & $+38.3$ \\
DeepSeek-V3           & 63.3\%$^{*}$ & 51.1\% & 27.2\% & 28.1\% & 29.6\% & $+21.5$ \\
Qwen-2.5-72B          & 57.0\%$^{*}$ & 42.7\% & 19.3\% & 20.0\% & 21.9\% & $+20.8$ \\
\hline
\end{tabular}
\end{table}

Previous work has shown that multiple-choice questions may not reflect the true ability of LLMs: options can be reached by elimination, back-solving from the answers, or distractor inference rather than by genuine reasoning and calculation \cite{zheng2024notrobust,li2024mcqreally,gupta2024answerorder}. Here we quantify this inflation exactly, on calculation questions from the domain. Using the identical 670-problem AtmosSci-Bench MCQ10 set, we grade every item in two ways: in option mode the model sees the four choices and returns a letter (the source benchmark's own protocol), and in code mode the options are hidden and the model must compute the answer as an executable \texttt{solve()} graded at 5\% relative tolerance (our protocol). Because both modes run over the same items under one harness, their difference isolates the effect of the answer format (Table~\ref{tab:mcq}). We evaluated three frontier models in both modes ourselves, and corroborated the result against the source benchmark's published option accuracies for three further models (Supplementary Table~S7).

Three alternative explanations must be eliminated before this gap can be read as a format effect. The first is a stricter numerical standard: we re-graded the same code answers at 10\% and 20\% relative tolerance, so that any near-miss is absorbed. The second is luck: we applied the standard chance correction for the four-way guessing baseline, $(p-\tfrac{1}{4})/(1-\tfrac{1}{4})$ \cite{lord1968statistical}. The raw gap on the full set is 30--45 points; applying both discounts together, option mode still stands 20--39 points higher (Table~\ref{tab:mcq}). Third, we tested whether defective answer keys explained the gap. Our audit flagged 19 of the 67 templates; removing these defective keys increased code-mode accuracy by 6--18 points but option-mode accuracy by only 2--6 points, showing that code mode is more sensitive to defective keys, whereas MCQ evaluation can still reward selection of the keyed option even when the key itself is problematic. After their removal, the chance-corrected gap remained 12--27 points (Supplementary Table~S7). In summary, on identical problems, requiring a model to compute rather than select removes between 12 and 39 points of measured accuracy that does not reflect computational ability, the width of the range reflecting both the model and whether the source benchmark's defective keys are included. The rescue mechanism is visible in the responses themselves: on problems whose code-mode answer is far wrong, the option-mode response often reproduces the same wrong value and then selects the nearest listed choice, in the clearest cases stating outright that the computed value is not among the options; and on multi-part problems the single graded letter credits responses whose executed code fails at least one asked quantity. One residual confound deserves note: code mode also requires writing a running program, a burden that is real for models that cannot code (the ClimateGPT reference models score higher under the prose protocol; Supplementary Table~S8). For configurations that can, the two protocols are near-tied on our core set (differences within about 3 points; Supplementary Table~S8), so the gaps in Table~\ref{tab:mcq} for the stronger models cannot be attributed to this burden; for the weakest entries the measured gap should be read as an upper bound on the format effect.

\subsection{Execution grounding saves cost and improves robustness}

\begin{figure}[t]
\centering
\includegraphics[width=\textwidth]{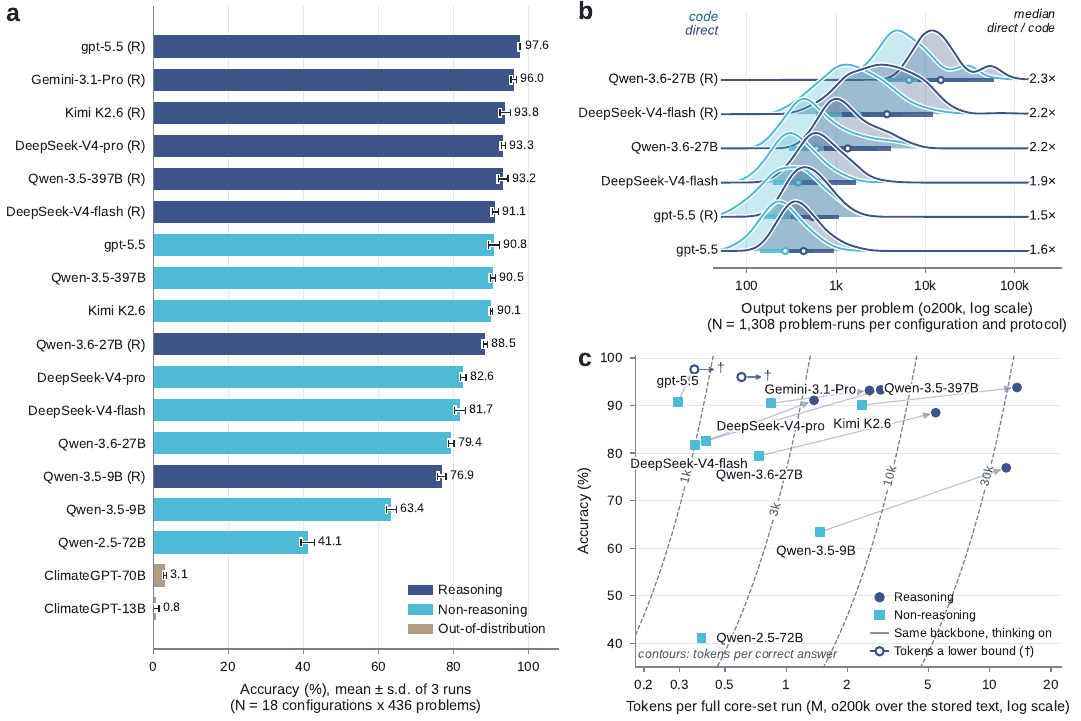}
\caption{\textbf{Overall model performance under execution-grounded evaluation.} \textbf{a}, Code-mode accuracy (mean $\pm$ s.d.\ over three runs) for the 16 leaderboard configurations, coloured by reasoning setting (labels ending in (R) denote reasoning enabled), together with the two domain-adapted ClimateGPT models shown separately as an out-of-distribution reference. \textbf{b}, Distribution of output tokens per problem under the two protocols, for the six configurations run under both, with the median direct-to-code ratio at the right. \textbf{c}, Accuracy against token cost; arrows join the non-reasoning and reasoning settings of the same backbone, dashed contours mark equal tokens per correct answer, and $\dagger$ marks the two configurations whose token counts are lower bounds because the endpoint returns only a summary of the reasoning.}
\label{fig:core_result}
\end{figure}

Figure~\ref{fig:core_result}a and Supplementary Table~S8 show the overall performance of 16 configurations of nine models (8 reasoning and 8 non-reasoning settings) and 2 fine-tuned domain models \cite{thulke2024climategpt} using the execution-grounded method. Frontier and reasoning models achieved substantially higher accuracy. Measured accuracy spans more than fifty percentage points, from 97.6\% for the strongest configuration (gpt-5.5 with reasoning) to 41.1\% for the weakest (Qwen-2.5-72B), confirming that the benchmark discriminates across the current capability range. The two domain-adapted ClimateGPT models score near zero (0.8--3.1\% in code mode), largely attributable to the weakness of their underlying base models; both score higher under the prose protocol (Supplementary Table~S8), the signature of a code-generation tax rather than a physics gap alone.

We compared all 436 problems under two modes: the code protocol asks models to write a \texttt{solve()} function and execute Python code to compute the final answer (our approach), whereas direct mode simply asks the model to reason by itself and return the result. Full protocol details are provided in Methods (Supplementary Figure~S1) and Supplementary Table~S5. Although code mode yields only modest accuracy gains, it substantially changes the reasoning process. In direct mode, complex tasks can require derivations long enough to exceed output limits; this occurred for three difficult problems. In code mode, even the longest, with 11 sub-questions, required fewer than 40 lines of Python and was solved correctly. Execution therefore decouples output length from computational depth, making complex multi-step tasks more reliable. A comparison of token usage is shown in Figure~\ref{fig:core_result}b.

For some tasks, code mode better ensures that the intended calculation is actually carried out rather than replaced by shortcuts \cite{zheng2024notrobust,yuan2024llms}. This is particularly important for iterative problems. For example, when particle diameter must be solved implicitly because the Cunningham slip correction depends on the unknown diameter, direct responses often stop after one substitution or merely state that iteration is needed, whereas code responses can execute the full iteration or root-finding procedure to convergence. Execution also reduces errors introduced during natural-language manipulation, such as lost signs or truncated expressions (Supplementary Tables~S10 and~S11), by evaluating the governing relation directly. Reasoning yields modest improvements in accuracy, with diminishing gains as model capability increases, but incurs a substantial increase in computational cost across all models (Figure~\ref{fig:core_result}c). Overall, the main limitation is often neither knowledge nor calculation itself, but the unreliability introduced by natural-language reasoning. Execution grounding therefore provides a more reliable and robust way to assess model calculation ability than direct evaluation.

We also made two types of variants for each task: numeric variants, in which the numbers are changed, and paraphrase variants, in which the numbers are fixed but the wording is altered (Methods, Supplementary Figure~S2, and Supplementary Table~S12). When the physics is held fixed while either the numbers or the wording is perturbed, measured accuracy remains almost unchanged: no configuration shows a significant gap after correction for multiple comparisons, and the leaderboard is preserved across the original, numeric, and paraphrase sets (Spearman $\rho=0.90$--$0.98$; Supplementary Figure~S4; Supplementary Tables~S13--S15). Robustness tracks capability. Because a numeric variant's reference answer is recomputed at freshly drawn inputs, a memorised textbook answer cannot transfer to it, and solving it requires computing the answer afresh; the absence of a measurable core-versus-variant gap is nonetheless reported as a bounded null (Methods), not as proof that no familiarity advantage exists. We also performed a prompt comparison, in which a functionally equivalent code-only prompt was set against the expert-persona prompt. The change has little effect on frontier models; its effect is concentrated in weaker models and operates primarily through executability rather than physical reasoning (Supplementary Tables~S16 and S17).

\subsection{Models silently misrepresent their execution process}

As stated in the Introduction, models can misrepresent their execution. Two prominent atmospheric benchmarks each contain documented examples of this issue. In AtmosSci-Bench \cite{li2025atmossci}, a ``code'' condition instructs the model to execute Python, but our re-examination found that no code was actually executed: the model returned an option letter while bypassing the calculation entirely, yet claimed to have performed it (Appendix~K of that work). In OneAtmos-Bench (Supplementary Table~S6 of that work) \cite{dai2025oneatmos}, a model was asked to determine the order of accuracy of an advection scheme. The task requires only a Taylor-series expansion of the symmetric six-point stencil, in which even-order derivative terms cancel, yielding the expected sixth-order accuracy. GPT-4o states that it performed a Taylor expansion and collected the error terms, but its visible argument instead relies on the heuristic that a wider stencil should imply higher order; the authors explicitly note that it did not calculate the coefficient cancellations. QwQ-32B similarly claims to have performed a numerical verification using a quadratic test function, without reporting the computation, and draws a convergence conclusion that such a test cannot establish.

We observed the same behaviour in our own runs. For the most severe form, complete fabrication of execution, we identified ten confirmed records covering eight distinct model--problem pairs, two of which fabricate identically in two independent runs; representative cases are reproduced in Supplementary Table~S18 and the complete set, with verbatim traces and replayed execution, is released with the benchmark. Flagged responses were screened by two automated annotators and verified by two domain experts. Notably, such fabrications were observed only in smaller models: eight records in ClimateGPT (70B and 13B) and two in Qwen-3.6-27B, one under the code protocol with reasoning enabled and one under the direct protocol without. We observed none in frontier models within the scope of this lexical, response-only scan. However, because none of these systems implements a mechanism to prevent or detect such fabrication, validation measures such as execution grounding which makes the computational process visible remain important. We also observed a milder form of execution misreporting, in which models claimed to apply one method or formulation while their actual calculations followed a different one. Detailed cases are provided in Supplementary Table~S19, with further discussion in the following sections.

\begin{figure}[t]
\centering
\makebox[\textwidth][l]{\sffamily\bfseries a}\\[1pt]
\includegraphics[width=0.97\textwidth]{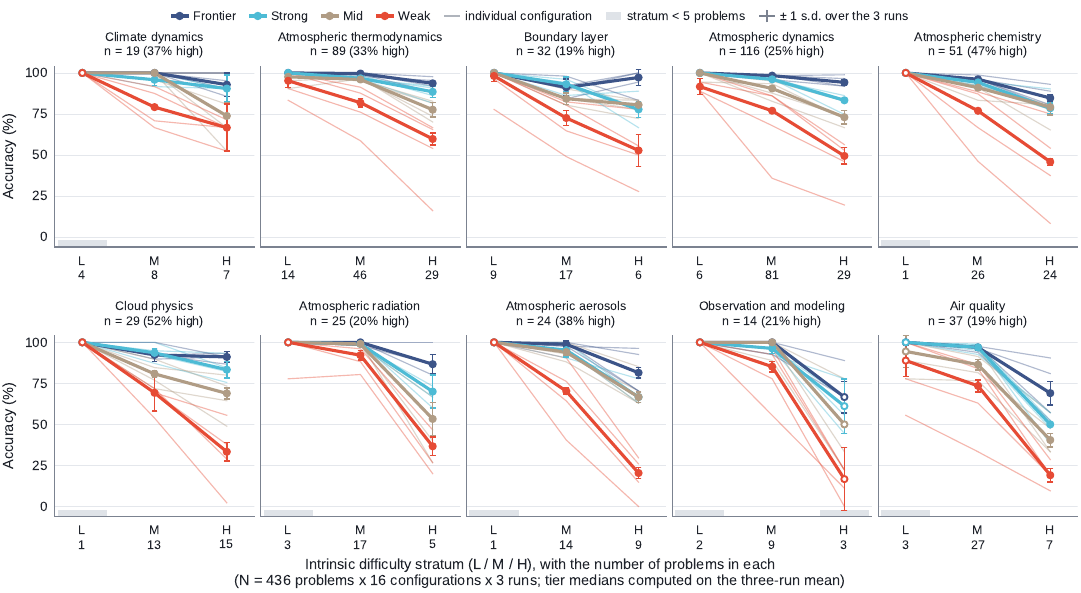}\\[4pt]
\makebox[\textwidth][l]{\sffamily\bfseries b}\\[1pt]
\includegraphics[width=0.97\textwidth]{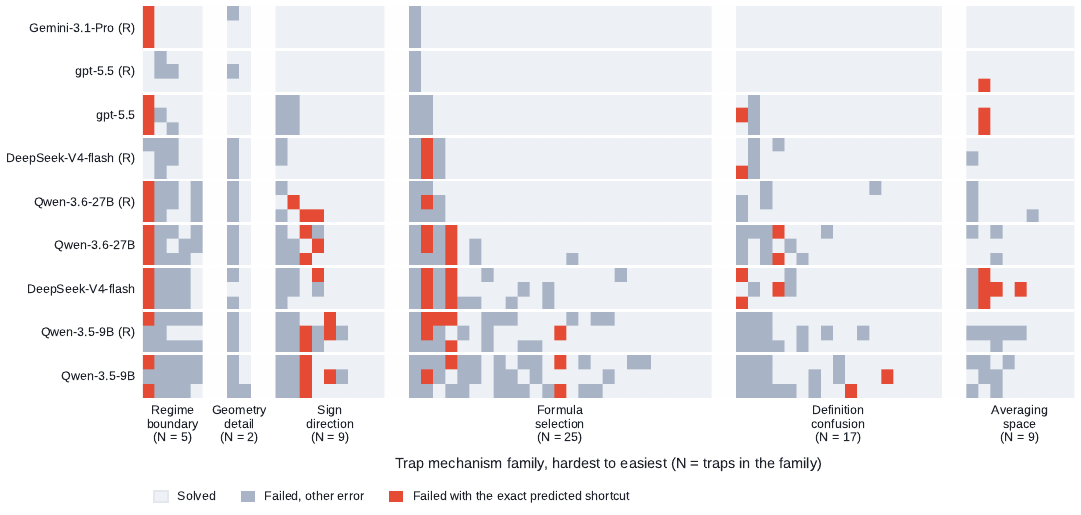}
\caption{\textbf{Performance variation across atmospheric domains and failure modes.} \textbf{a}, Accuracy across atmospheric subfields and difficulty levels. \textbf{b}, Per-trap outcomes for each configuration across the six trap mechanism families. Red denotes failures returning the exact shortcut value predicted during trap construction, whereas grey denotes other failures; pale grey marks a solved trap.}
\label{fig:category_and_trap}
\end{figure}

\subsection{Performance across subfields}

We also mapped computational ability across different domains of atmospheric analysis (Figure~\ref{fig:category_and_trap}a; Supplementary Table~S20). Performance is higher and more consistent across difficulty levels in subfields such as climate dynamics, thermodynamics, boundary-layer meteorology or dynamics, but lower and more differentiated in categories such as chemistry, air quality, aerosols, and cloud physics (lowest, 76.1\%). Performance differs little across categories on low-difficulty questions. A category's share of high-difficulty problems accounts for part of its ranking (accuracy declines as this share rises, $r=-0.49$; Supplementary Figure~S3), but difficulty mix does not exhaust the effect: much of the remaining variation tracks task-type composition rather than domain. Lower-performing categories contain a higher proportion of tasks requiring iteration or root-finding, involving exponential or logarithmic relations, or requiring the computation of multiple quantities, regardless of whether the tasks are of medium or high difficulty.

Manual inspection of the hardest problems revealed four recurring archetypes. The first is a long multiplicative chain involving heterogeneous units, in which each step depends on the preceding one. Models tend to mix units, lose symbols or parameters, or propagate incorrect intermediate results, so an error at one step compounds errors in subsequent steps (Supplementary Note~S1, Example~1). A related problem is conservation bookkeeping: models may correctly write individual reactions but fail to maintain constraints across the full network, such as elemental conservation, mass balance and molar balance (Supplementary Note~S1, Example~2). In these cases, models often know the correct formulas or rules but do not consistently follow them throughout the derivation, resembling execution misrepresentation. Although both failures arise from a broken deduction chain, conservation-bookkeeping errors can occur whenever multiple species or reservoirs must be tracked, even when the chain itself is not long. The third archetype is inappropriate formula selection: models may apply an empirical formula outside its valid regime, misremember its coefficients, or confuse nearby empirical relations (Supplementary Note~S1, Example~3). The last is a knowledge or interpretation issue: Models confuse number and mass/weight distributions, radius and diameter, or the statistical moments required to convert between distribution representations; they may also calculate a related quantity rather than the one requested, such as returning an amplitude when the problem specifies a peak-to-peak variation (Supplementary Note~S1, Example~4). The iteration-truncation issue introduced in Section~2.3 was also observed, particularly in tasks requiring iterative solutions in cloud physics and atmospheric aerosols. 

Across categories, model performance is related to two main factors: how ``template-able'' the underlying physics is and how much domain-specific judgement is required. Models are strongest when retrieving a single well-known equation is sufficient, and weakest when solving the problem requires regime judgement, step-by-step deduction, sustained multi-step bookkeeping, or expert knowledge to select the appropriate parameterization or concept. We also ran the identical construction pipeline on four non-atmospheric environmental domains (hydrology, environmental chemistry, ecology and biogeochemistry, and soil mechanics) and observed the same capability ordering and discrimination (Supplementary Figure~S5; Supplementary Table~S21).

\subsection{Trap tasks expose template matching over physical reasoning}

To test whether measured accuracy reflects genuine physical reasoning or the reproduction of a memorised solution template, we built trap tasks by altering a single condition of a parent problem so that the parent's canonical solution method no longer applies (Methods; Supplementary Tables~S22--S23). The traps are grouped into six types: regime-boundary violations (standard approximations pushed outside their valid range), overlooked geometry, sign and direction errors, formula selection, definition confusion between similar parameters, and inappropriate averaging. Regime-boundary traps proved to be the hardest, while inappropriate averaging was the easiest. The results showed that the weaker the model, the more vulnerable they are to traps: the failure rate on traps whose parent the model solves falls from 36\% to 3\% across the nine configurations (Supplementary Table~S23). Enabling reasoning lowers the gap for every backbone but cannot eliminate it, and it still did not prompt the model to ask whether the canonical method still applies. Figure~\ref{fig:category_and_trap}b resolves this by trap and by configuration. Red marks a failure whose returned values match the shortcut predicted when the trap was built, and grey a failure for any other reason.

The failures are systematic. Several models independently return the exact numerical vectors predicted by our advance shortcut analysis. The underlying mechanisms are also similar across models: retaining an approximation beyond its valid regime, dropping spherical metric terms, carrying over a sign convention after the direction changes, reporting a nearby quantity in the wrong reference frame, and averaging in the wrong physical space (Supplementary Note~S2, Examples~1--5). Models struggle most with regime judgement: regime-boundary traps have the lowest pooled solve rate of the six families (42\%), and even the strongest configurations are captured by the clearest case, whose exact predicted shortcut is returned by seven of the nine configurations (Supplementary Tables~S22--S23). Geometric details constitute another difficult class, where models overlook physical or geometric constraints and therefore select incorrect parameters or formulas. We also observe execution misrepresentation: a model may correctly identify a parameter as a layer average in its comments or definitions, yet use it in the calculation as though it were a surface flux (Supplementary Note~S2, Example~5). Other failures include losing signs, confusing related concepts, and applying averages when the problem instead asks for values under specific conditions. These failures are consistent with the broader patterns described above and are examined further in the next section.

\subsection{Frontier models remain limited by domain-specific judgement}

\begin{figure}[H]
\centering
\includegraphics[width=\textwidth]{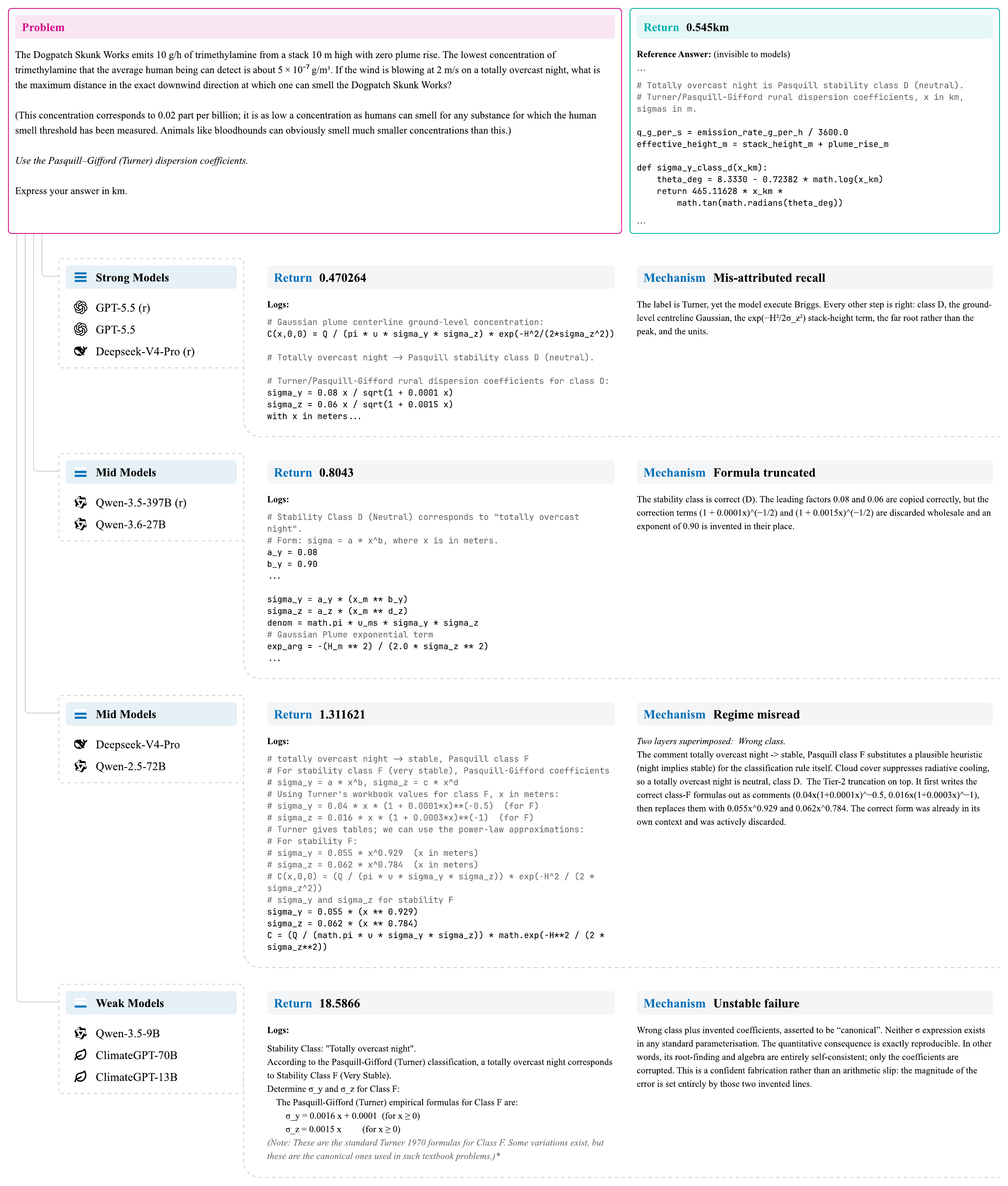}
\caption{\textbf{Failure modes on the Dogpatch Skunk Works Gaussian-plume task.} No configuration solves it, and the failures differ in kind: strong models misreport the parameterisation used, intermediate models misclassify regimes or alter formulas, and weaker models show unstable sign, unit, and arithmetic errors.}
\label{fig:gaussian}
\end{figure}

The diagnostics above describe what current models, even with execution grounding, still cannot reliably do. Strong reasoning models achieve satisfactory scores on most tasks, yet remain vulnerable to certain difficult cases. We identified two recurring patterns in frontier models. First, LLM performance is strongly related to how readily the underlying physics can be reduced to a canonical closed-form method: models remain highly inclined to apply familiar empirical formulas or solution templates while overlooking task-specific conditions. Second, models are easily misled by traps that alter the initial conditions or physical setting of a task. When a stated condition invalidates a formula's physical assumptions, violates its geographical applicability, falls outside its regime of validity, or is itself physically impossible, models often apply the original formula or method unchanged and ignore the alteration entirely. These results suggest that strong LLMs are limited less by calculation than by judgement: they fail to recognise when a canonical solution no longer applies and when a task-specific treatment is required. This is precisely where professional knowledge and judgement remain essential and, at least for now, cannot be reliably replaced by LLMs. After excluding problematic benchmark items, the remaining hard cases showed either cross-model convergence on the same wrong method or highly scattered failures, revealing shared methodological biases or the absence of a stable solution strategy.

For most models, several additional failure modes recur. LLM performance is highly unstable in long-chain deduction: when tasks require many sequential steps, models tend to truncate derivations or iterations, lose signs or intermediate results, and fail to maintain constraints across a full network even when they can state those constraints correctly. They also confuse related but distinct physical quantities, mix up units and signs, conflate similar concepts, and take averages where values under specific conditions are required. An instructive case is the Dogpatch Skunk Works emission dispersion task, which requires a Gaussian plume model. The problem specifies a ``totally overcast night'' and, to remove ambiguity, explicitly requires the Pasquill--Gifford (Turner) dispersion coefficients. No configuration solves it on a majority of its three runs (2 of the 48 measurements succeed), and the failure modes differ across capability levels. The case is reported as a qualitative illustration and is excluded from the quantitative ceiling statistics (Methods). Strong models such as gpt-5.5 (reasoning) computed $\sigma_y$ and $\sigma_z$ from the Briggs parameterisation while claiming to have used Turner: the stated method and the executed parameters disagree, a misreport that makes the answer merely ``look'' compliant. But these strong models were correct on the remaining solution chain. Weaker models such as Qwen-3.5-397B-r and DeepSeek-V4-pro instead struggled with the long formula or conditions: the Briggs class-D parameterisation is $\sigma_y = 0.08\,x\,(1 + 0.0001x)^{-1/2}$, yet Qwen-3.5-397B-r computed $0.08\,x^{0.90}$, copying the leading factor, discarding the correction term, and inventing an exponent of $0.90$ in place of the actual calculation. DeepSeek-V4-pro, on the other hand, failed at an earlier step: it assigned stability class F (stable), reasoning that a nighttime case must be stable. This overlooks the fact that cloud cover suppresses nighttime radiative cooling, so a completely overcast night is neutral (class D) rather than stable, exactly as Pasquill's original scheme prescribes (nighttime with more than 4/8 low cloud is class D regardless of wind speed). The model thus applied a plausible heuristic (night implies stable) in place of the classification rule the problem's conditions actually invoke. The weakest models fail unstably rather than systematically: dropped signs, unit mistakes, magnitude errors, and fabricated arithmetic occur essentially at random, producing errors of unpredictable magnitude, from a factor of a few to several orders of magnitude. This single task thus surfaces the full spectrum of failure modes an LLM can exhibit, and, most strikingly, reveals that even strong models may claim to use one parameterisation while actually executing another, misreporting their own method to make the answer appear compliant. The full task and representative model outputs are given in Figure~\ref{fig:gaussian}.

\section{Discussion}\label{sec:discussion}

In this work, we developed AtmosCoder-Bench, an LLM benchmark whose innovation lies in two aspects. First, its task database is strictly constructed. Unlike previous work, we rigorously cleaned the question set to ensure that no unintended ambiguity remains, that every question is solvable by domain experts, and that each question has a unique, verifiable solution, while no solution procedures or give-away formulas are embedded in the question itself. The construction is semi-automated: we programmatically extract QA pairs from representative textbooks and review the answers, and for tasks without a textbook answer, three independent LLMs must agree on the answer, followed by a manual check. This yields a task database that is cleaner, more reliable, professionally grounded, and uniquely solvable, with construction and verification largely automated so that human labour is reduced and the benchmark can be scaled to more questions at low cost.

Second, we evaluated LLMs under an execution-grounded architecture. Each model is required to write a \texttt{solve()} function and execute it in Python, so whether to use code or direct calculation is no longer the model's choice. Compared with free-form direct calculation, execution grounding makes the computational process visible and auditable at a lower token cost, where reasoning only spends $1.3$--$2.1\times$ as many tokens per full run, while maintaining comparable accuracy. We also found that, under execution grounding, strong models are highly robust to both numerical and paraphrase perturbations. Together, these results show that execution grounding provides a more stable and reliable way to assess LLM calculation performance.

With this benchmark design, we reached several findings on LLM quantitative analysis in atmospheric and broader environmental domains. First, we quantified the inflation introduced by multiple-choice evaluation, finding that it overstates apparent LLM calculation ability by 12--39 percentage points, depending on the model and on whether the source benchmark's defective answer keys are included. Second, most models are systematically weak on tasks requiring long-chain computation with heterogeneous units or domain-specific reasoning. They truncate derivations or iterations, lose signs and intermediate results, fail to maintain constraints across multiple steps, and confuse closely related quantities or concepts. In many cases, they can state the relevant rules correctly but fail to apply them consistently throughout the reasoning process. Therefore, LLM calculations on domain-specific tasks cannot be fully trusted, not primarily because of missing knowledge, but because models often fail to follow the knowledge they possess consistently and therefore do not reason as reliably as a human specialist.

Finally, our design reveals the true calculation ceiling that persists even in state-of-the-art models. Across categories and trap tasks, we found that frontier LLMs are still weak when problems require task-specific judgement: they often fail to check whether the stated conditions remain within the valid regime of a canonical method. When we altered the initial physical setting so that the original method no longer applied, models frequently failed to recognise the mismatch and instead reused the canonical formula while ignoring the task-specific conditions, even when explicitly instructed to account for them. In the Gaussian plume case, even though the task explicitly required the Pasquill--Gifford (Turner) dispersion coefficients, strong models such as gpt-5.5 (reasoning) still used the Briggs parameters while claiming to have used Turner, so that the answer merely appeared compliant. This constitutes the judgement ceiling of LLM calculation in the environmental domain: frontier models may possess strong reasoning and calculation abilities, yet they lack the professional judgement required by a specific task and cannot fully resist applying common, popular solution patterns while ignoring the task's particular setup. Together with the failures described above, this marks a key area where expert judgement remains essential and cannot yet be reliably replaced by LLMs.

Several limitations should also be considered. First, the choice of tasks: as a consequence of our automated construction and the requirement that every question be computable, we retained only tasks that can be independently verified and executed. In particular, we removed tasks with open-ended or multiple valid answers, although many real quantitative problems in the field are of this kind. The database is therefore verifiability-filtered: it cannot cover the full range of problems in the domain, not to mention that self-contained textbook computation does not represent the full breadth of quantitative work in the field. Second, for tasks without a textbook written answer, we used the consensus of three LLMs as the reference. Consensus does not guarantee correctness, as demonstrated by several of our case studies (such as the trap cases). We mitigated this with two mechanisms: all generated answers were manually verified, and every case analysed in detail either has a textbook answer or was re-examined by more than one human expert. This, however, requires additional human effort and prevents the pipeline from being fully automated. Third, we explored only 131 tasks from 4 other fields beyond atmospheric science; although these yielded similar results, a firm conclusion for the general environmental domain would require more tasks covering a broader range of subfields.

Our work provides a clean, fully verifiable benchmark that enables a more accurate evaluation of LLM performance on quantitative tasks and implemented in atmospheric analysis domain. Through execution grounding, we successfully separated the evaluation of genuine reasoning from mere answer generation, revealing several important findings about current LLM capabilities in this area. Nevertheless, for the reasons above, the benchmark cannot be regarded as covering all aspects of the domain. Future work should be more application-grounded: rather than drawing tasks from textbooks, it should target problems encountered in real-world environmental management, such as emission analysis and pollution alerting. We encourage such exploration and believe our work lays a foundation for studying LLM applications in this and related domains.

\section{Methods}\label{sec:methods}

\subsection{Benchmark construction}
AtmosCoder-Bench comprises 436 self-contained computational problems drawn from 13 established
textbooks spanning undergraduate to graduate level across the atmospheric and environmental sciences
(Supplementary Table~S4), each labelled with one of ten physical categories and one of three
difficulty levels (Table~\ref{tab:stats}; category definitions and the difficulty rubric in
Supplementary Table~S6). The 436 problems carry 734 graded sub-answers: 298 ask for a single quantity
and 138 for two or more, up to fourteen. Items enter through the staged admission pipeline of
Fig.~\ref{fig:pipeline}, whose defining property is that every reference answer is
execution-grounded---it is the return value of an executable reference solver---so that the released
corpus re-verifies end to end, every solver re-executed against its stored answer, without a single
model call.

From OCR'd textbook sources we extract candidate problems that are self-contained and numerically
answerable, storing each with its verbatim statement and source, and discard derivation-only,
figure-dependent and external-data-dependent items along with any that are ill-posed, ambiguous or
admit multiple defensible answers.

Every retained problem is paired with a reference \texttt{solve()} function generated \emph{blind}: the
model that writes it sees only the problem text, never the target answer. The solver obeys a fixed
contract, reproduced verbatim in Supplementary Table~S5: every given quantity becomes a parameter
carrying its stated value as a default, only the Python standard library may be used, all unit
conversions are explicit, and the return value is keyed by sub-answer id in the order asked, each entry
holding a number and a unit string. Code runs in an isolated subprocess under a hard timeout, and only
solvers that execute and return the contracted structure are retained.

A reference value is admitted only when independent derivations by models from three different
developers agree, sub-question by sub-question: matched against the published textbook answer where a
solutions manual exists, and frozen by cross-developer consensus otherwise. Disagreement with the book
exposes misprinted keys, corrected and logged as errata; disagreement among the models exposes damaged
extractions, repaired or discarded.

Because a correct final value does not establish that a computation was performed, every solver then
passes a deterministic audit that rejects answer values baked in as literals, and inputs the solver
ignores, by abstract-syntax-tree inspection and input perturbation (Supplementary Table~S24). Structural
checks and a three-model consensus review then screen for keys that collapse under grading, question--answer
misalignment and items unsolvable as stated.

A de-scaffolding pass strips supplied formulas and recallable constants from the statement so that
each problem tests method rather than substitution. Each edit is retained only under a triple
gate---the answer must still be recovered from the stripped text, an independent audit must confirm
that the item-specific data is intact, and a second-developer model must reproduce it---and any edit
that introduces ambiguity is reverted. The de-scaffolded statements are the released problem texts.
What this pass buys is measured directly by an ablation: four models spanning the capability range
were run on both versions of the 169 affected problems under the code protocol, fully paired. The
strongest model loses 1.8 points when the scaffolding is removed, whereas a 9-billion-parameter
model loses 18.5, a monotone gradient in scale that widens the spread across the four models from
10.3 to 27.0 points; the released statements therefore test knowledge the textbook phrasing had
handed over, and leaving the scaffolding in place would over-credit smaller models most
(Supplementary Table~S27).

Each admitted item is finally extended into perturbed variants that preserve computational structure
(Supplementary Table~S12). Physical constants are moved out of the solver signature before any
perturbation, with the output proven bit-identical before and after, so that variant generation cannot
alter a constant even by accident. Of the 436 problems, 346 admit safe
perturbation and carry five numeric variants each (1{,}730 in total), with ground truth recomputed at
the perturbed point by a full-precision twin of the reference solver freed of the textbook's own
rounding; the remaining 90 are retained core-only because re-sampling their inputs would not yield a
well-posed twin. Every problem additionally receives five paraphrase variants (2{,}180) that reword the
statement with values and answer unchanged, admitted only if every stated magnitude survives the
rewrite. All variants pass the same faithfulness audit and a cross-developer check by the two
developers other than the author, giving an evaluated corpus of 4{,}346 instances carrying 7{,}029 graded quantities.

\subsection{Human adjudication}
Large language models are used throughout construction only as \emph{proposers}---of candidate
solvers, answers and rewordings---while admission is decided by criteria that do not require trusting
any single model: deterministic execution, anchoring to the textbook's printed answer, agreement
across developers, and adjudication by a domain expert. Automated auditors over-flag, so no rule-based
check or consensus review ever decided an outcome on its own; each raised a suspect, an expert chose
between repair, regeneration and removal, and any repaired artefact re-entered the verification gates
from the start. Beyond that, an expert re-derived the agreed computation for every admitted problem
before its value was frozen, and read the statement, reference solver and stored answer of all 436
problems in final de-scaffolded form before release. The physics content itself is not model-authored:
every problem is taken from a published textbook. Every decision point, and what was decided at each,
is enumerated in Supplementary Table~S3; the four models used as construction tools, with their exact
API identifiers and roles, are in Supplementary Table~S2; and the yield of the deterministic
faithfulness audit, which is the evidence that these gates bind rather than decorate, is in
Supplementary Table~S24.

\subsection{Trap diagnostic}
Beyond the graded corpus we build a held-out diagnostic of 67 \emph{traps}, one per distinct parent
core problem, spanning all ten categories and six mechanism families. A trap is a minimal,
single-trigger perturbation of a certified core problem: one detail is changed so that the problem's
own canonical, reflexively applied method becomes a specifically wrong shortcut, while the physically
correct method yields a different answer. Candidates are either mined from stored results, where several
independent strong models already fail on a parent with the same wrong value, or authored by altering a
regime, definition or sign and recomputing the correct answer. Each released record carries a
construction note identifying its parent, trigger and predicted shortcut; 62 of the 67 traps were
proposed by GPT-5.5 and adjudicated by a model from a second developer. Each is admitted only after checks
beyond standard answer certification: exactly one
defensible answer; a separation check, the shortcut solver's output differing from ground truth by far
more than the grading tolerance (13--880\%, median 65\%) so that falling for it is always visible to the
grader; a fairness check, everything needed being stated in the problem and a carefully reasoning
frontier model recovering the answer, so that a trap is a diagnostic rather than a test of ambiguity;
and a self-containment audit. Every surviving candidate was then reviewed by hand for physical
correctness. Because the proposing model is itself an evaluated subject and the fairness gate
conditions admission on a frontier configuration recovering the correct answer, trap accuracy at the
top of the capability range is favoured by construction and the small frontier Trap Gap is best read
as a lower bound on shortcut susceptibility; the capability and reasoning gradients, by contrast, are
reproduced within model families that played no part in trap construction.

Each record stores the predicted shortcut as an executable solver together with its full output
vector, so that falling into the trap can be distinguished from failing for unrelated reasons: a run
counts as \emph{captured} only when every returned sub-answer matches the shortcut vector within 2\%.
The unperturbed parent is the built-in control, and the whole set is held out from the main corpus,
leaving leaderboard denominators unaffected. Nine configurations were evaluated, three runs each.
Mechanism-family counts and per-trap constructions are in Supplementary Table~S22, and overall trap
performance in Supplementary Table~S23.

\subsection{Comparison against an external multiple-choice benchmark}
Comparing answer formats requires content that neither protocol was designed around, so the
comparison was carried out on a published third-party benchmark rather than on our own corpus. We used
AtmosSci-Bench \cite{li2025atmossci} in its official MCQ10 configuration: 67 problem templates, each
instantiated at 10 symbolic perturbations of its inputs, giving $N=670$ instances. Statements, options
and answer keys were taken as published and converted only in representation, each key parsed into the
value-and-unit sub-answer form our grader consumes; one template asks for two flow-type classifications
alongside its numeric quantities and is graded on the numeric quantities only. Every model is compared
across the two modes on the identical 670 instances: in \emph{option} mode the source benchmark's own
protocol was used unchanged, the model seeing the four options and being graded on the letter it boxes;
in \emph{code} mode the option list is withheld and the executed output of the returned
\texttt{solve()} is graded numerically. The comparison is therefore paired at the instance level and
differs only in what the model is asked to produce.

Two properties of the design guard the interpretation. The option-mode numbers are not produced by our
grader, so an inflated format gap cannot be an artefact of our scoring: for three models we ran the
source harness ourselves and for three others we quote the option-mode accuracies published by the
AtmosSci-Bench authors on the same set, reporting the two groups separately. And because a strict
numeric threshold could itself depress code-mode accuracy, every code-mode run was re-graded offline at
5\%, 10\% and 20\% relative tolerance (Supplementary Table~S7).

A numeric grader is only as sound as the keys it grades against, so the imported set was audited with
the same machinery, and under the same conservative admission rule, used to certify our own corpus: a
template was condemned only where two independent auditor models both flagged it \emph{and} two blind
solver models from different developers either agreed with each other against the key or were mutually
inconsistent. That intersection flags 21 of the 67 templates; a final unit-aware pass---the specificity
control for the procedure as a whole---retracts two whose keys are correct but expressed in a
commensurate unit, so the audit condemns 19 templates (190 instances), nine of them carrying keys that
both blind solvers contradict.

The defective instances were deliberately retained rather than removed: the published option-mode
accuracies and our own option-mode runs are all computed over the full MCQ10, so discarding problems on
our side would break every cross-protocol and cross-paper comparison. All results are therefore
reported twice---over the full 670 instances for comparability with the published numbers, and over the
480 instances from unflagged templates for a defect-free measurement. One configuration,
DeepSeek-V4-flash with reasoning, is excluded from both protocols here because its option-mode arm
returned no letter on a large fraction of items; dropping its code-mode arm too keeps the pairing
symmetric (Supplementary Table~S7). Restricting both protocols to the unflagged templates raises
code-mode accuracy by 6--18 points but option-mode accuracy by only 2--6, and the chance-corrected gap
settles at 12--27 points: defective keys inflate the full-set gap, by penalising the arm that computes
and not the arm that selects, but they do not create it.

\subsection{Cross-domain generalization}
To test whether the construction method rather than the subject matter produces a discriminative
benchmark, the identical pipeline was applied to four non-atmospheric environmental domains:
hydrology (37 problems), environmental chemistry (21), ecology and biogeochemistry (19) and soil
mechanics (54), 131 problems in total, drawn from openly licensed university course materials and one
geotechnical engineering textbook. Ground truth was admitted only on position-by-position agreement,
within tolerance, between the authoring model and two independent witnesses from different developers,
followed by the same anti-hardcode audit. Where a source ships an official answer key, stored answers
were cross-checked against it and every mismatch adjudicated by hand, separating unit and rounding
conventions from genuine disagreement; problems whose official answer proved unreachable from the
statement alone were removed. All 131 released solvers reproduce their stored answers under machine
verification. Five non-reasoning configurations overlapping the core leaderboard were evaluated under
the code protocol, one run per model.

\subsection{Evaluation protocols}
Models answer under one of two protocols that share an identical system message (``an expert in
atmospheric science'') and differ only in who executes the arithmetic. In \emph{code} mode the model
receives only the problem text---no formulas beyond those intrinsic to the statement, no expected
answer---and returns a \texttt{solve()} function under the same contract as the reference solvers,
which is executed in an isolated subprocess under a hard 10\,s timeout; the executed output, never the
prose, is graded. In \emph{direct} mode the model reasons in prose and reports one value per asked
quantity as a boxed number with its unit; declaring the unit gives the grader the same
unit-reconciliation ability as in code mode. The direct protocol was run on a
six-configuration subset spanning the capability range and both reasoning settings, together with the
two out-of-distribution climate models, so that every code-versus-direct comparison is paired at the
configuration level on the identical 436 problems.

An ungradable \emph{content} response---code that will not run or is absent, or a direct answer with no
boxed value---is fed back verbatim with the execution error or a format reminder, for up to five
attempts; output that remains ungradable is scored as a failure, so repair fixes format, never physics.
A wrong but gradable answer is final and is never retried. Transient API failures are retried without
counting and, if they never clear, the item leaves the denominator, so accuracy is
$\mathrm{passed}/(\mathrm{passed}+\mathrm{failed})$. One class of non-completion is scored
as a failure rather than excluded: under the direct protocol, seven records for gpt-5.5 with reasoning
returned no answer because the in-band derivation exhausted the serving limit on the three
computationally deepest problems. Since the same model solves those problems under the code protocol,
the non-completion is attributable to the protocol rather than to infrastructure, and these records are
counted as failures and flagged in the released files.

Every configuration is run three independent times with the sampling seed offset per run, and results
are reported as the mean and sample standard deviation over runs. The reasoning-permissive code prompt
is the primary setting; a functionally equivalent code-only variant, differing in the system persona
and in whether prose reasoning is permitted before the code block, is used only for the
prompt-sensitivity analysis (Supplementary Table~S16). Tokens are counted uniformly with one tokenizer
(\texttt{tiktoken o200k\_base}) over the stored text rather than each provider's billed usage, so costs
are comparable across models.

\subsection{Grading}
Grading is answer-keyed, unit-aware and identical across protocols. Let the key for a problem list
asked quantities $q=1,\dots,Q$, each with a set of accepted values $E_q$, where multiple accepted
values encode legitimate sign or convention alternatives. A returned value $a_q$ passes if
\begin{equation}
\min_{e \in E_q} \frac{|a_q - e|}{|e|} \;\le\; \tau, \qquad \tau = 0.05,
\end{equation}
where zero-valued references are compared absolutely ($|a_q|\le\tau$) and signed infinities are
matched symbolically. The comparison is evaluated after unit reconciliation: when both sides declare
units, values are converted to a canonical form using a fixed factor table that covers compound units
and offset scales such as Celsius and Kelvin, so that a correct answer expressed in a commensurate
unit is not penalised. A problem passes only if every asked quantity passes. The remaining
conventions---positional fallback when a model keys its answers differently from the stored labels,
tolerance of loosely formatted numbers, and the offline re-grading that establishes the 5\% operating
point as a plateau rather than a tuned threshold---are implemented in the released grading module
and summarised with the prompts in Supplementary Table~S5; the size of the tolerance effect is
measured, on the external multiple-choice set, in Supplementary Table~S7.

\subsection{Fabricated-execution audit}
Fabricated execution is detected by scanning every stored response, across the eight configurations run
under both protocols, for assertions that a machine ran a computation and returned a result. Detection
is lexical and therefore a lower bound, and reasoning traces are outside the scope by design, so the
audit establishes that the behaviour occurs and is reproducible rather than estimating how often it
occurs. Adjudication then proceeds in two stages. Each flagged response is first judged against a
written criterion---fabricated where a completed run is asserted as accomplished fact, honest where the
text hedges, states an intention, hand-derives, or merely describes the return structure---by two
independent automated annotators, whose verdicts agree on 25 of the 26 flagged responses (Cohen's
$\kappa=0.92$). Two domain experts then review every admitted record in full: the complete response,
the code the harness executed, its replayed output and the graded verdict. All ten admitted records
were confirmed as fabricated. The single record on which the automated annotators disagreed was judged
honest and is not counted, because the value it mentions is printed in the question itself and the
surrounding passage plans the program rather than reporting a run. Since every confirmed record carries
the same label, $\kappa$ is undefined for this stage and agreement is reported instead, 10 of 10.

\subsection{Statistical analysis}
Problem-level outcomes are stabilised with a majority-of-three rule across runs, and a parent's
variant family counts as held when at least three of its five variants are solved. Contamination and
linguistic robustness are assessed on fully paired parent subsets (346 parents for numeric, 436 for
paraphrase): for each model the paired shift $\Delta$ between core and variant accuracy is tested with
an exact two-sided McNemar test on discordant pairs, with 95\% confidence intervals from the analytic
paired standard error, a seeded 2{,}000-resample bootstrap giving concordant intervals, and Holm
correction across the sixteen per-model tests within each family. Leaderboard stability across sets is
quantified with Spearman and Kendall rank correlations, and single proportions carry Wilson score
intervals. As a mechanism-level check, failed variant answers are tested for being
literal echoes of the parent's answer on sub-answers that verifiably discriminate between the two
(Supplementary Table~S15), and a problem is reported as leaked only when at least two independent models
echo it while also solving the parent. The paired design bounds a null result rather than proving
absence: the analytic 95\% intervals on the per-model shift are $\pm1.0$ to $\pm4.2$ points wide
(median $\pm2.0$), so a residual familiarity advantage below roughly two points would escape detection
at this sample size. The echo test identifies 12 problems as leaked, and excluding all 32 flagged
problems leaves the largest per-model shift essentially unchanged ($+2.3$ to $+2.2$ points). Reliability is summarised as pass@3 and all@3 (solved in at least one, and in
all three, of the runs). For the trap diagnostic the headline statistic is the Trap Gap, the
failure rate on traps restricted to those whose parent the same model solves in the run-matched core
experiment,
\begin{equation}
\mathrm{TG} \;=\; \frac{1}{R}\sum_{r=1}^{R}
\frac{\bigl|\{\,i \in S_r : \text{trap } i \text{ failed in run } r\,\}\bigr|}{\bigl|S_r\bigr|},
\qquad S_r = \{\, i : \text{parent of } i \text{ solved in run } r \,\},
\end{equation}
which conditions out general incompetence and isolates susceptibility to the trigger.

\subsection{Residual-failure screening}
Problems analysed as evidence of a capability ceiling are screened first, so that a failure is
attributed to models rather than to the corpus. Starting from the lowest-passing core problems under
the code protocol, a candidate is admitted only if three tests hold. First, the reference values must
not reappear under any permutation, sign flip or unit rescaling of the model's returned values---the
defect classes behind our earlier dataset repairs---so that an answer-alignment artefact cannot
masquerade as a reasoning failure. Second, the reference must be independently re-derivable in closed
form, or convergently reached by several frontier configurations, over and above the corpus's standing
cross-developer certification. Third, the model--reference disagreement must trace to neither an
interpretive reading of the question nor variance among published empirical fits of the same chart.
Candidates failing any test are excluded and released with reasons (Supplementary Table~S25), so the
lowest-scoring items are not automatically the most informative about model ability; the three that
survive the screen, with the answer clusters into which the 48 measurements fall, are in Supplementary
Table~S26. Those three pass 8, 9 and 12 of their 48 measurements (16 configurations, three runs), and
their failure signatures divide. On two of them the field fails in consensus---27 runs return the same
wrong value on the effective-gravity problem, and 21 drop the same prefactor term on the
saturation-vapour-density problem---so a cross-model majority vote would canonicalise the error, which
bounds what consensus-based answer verification can certify and is one reason our ground truth is
anchored to the textbook's printed answer wherever one exists. The third inverts the signature: its
twelve passes come from the strongest configurations, while the 36 failures scatter across more than
forty orders of magnitude without a single repeated non-zero value. The screen governs which failures are admitted as \emph{quantitative} evidence of a
capability ceiling. It does not govern the worked case study of the Results, which is chosen for the
range of failure mechanisms a single task exposes rather than for the attributability of its score, and
which is therefore reported as a qualitative illustration and excluded from the ceiling statistics.

\subsection{Models evaluated}
We evaluate sixteen configurations under the code protocol, spanning nine backbones from five
developers, seven of them with reasoning both enabled and disabled; the direct protocol is run on a
representative six-configuration subset. Two domain-adapted models (ClimateGPT-13B and -70B) are
evaluated under both protocols as an out-of-distribution reference rather than as leaderboard entries,
and two further models only on the external multiple-choice set, where their option-mode accuracies are
quoted from the source publication. Reasoning is toggled explicitly for reproducibility, using a
provider-specific thinking flag or a reasoning-effort setting; non-reasoning runs use low-temperature
decoding, and a 16k-token output budget is used throughout. Exact API identifiers, access routes and
settings for every configuration are in Supplementary Table~S1. Two evaluated models also served as
blind verification solvers during construction. This overlap is disclosed rather than hidden and is
bounded by three properties: most problems are anchored to the textbook's printed answer rather than to
model agreement, the evaluated statements are the de-scaffolded rewrites rather than the texts the
verifiers saw, and the numeric-variant family probes for any residual familiarity advantage. On the
cross-model route, however, admission required those models to solve a problem at certification time,
and we state that one-way selection as a limitation.

\backmatter

\bmhead{Supplementary information}
Supplementary Information (Tables S1--S27, Figures S1--S5, Notes S1--S2) follows the references at the end of this preprint.

\bmhead{Acknowledgements}
We thank external experts from governmental environmental management agencies in China for
their expert review and domain guidance.

\section*{Declarations}

\begin{itemize}

\item \textbf{Funding.} This work was financially supported by the Natural Science Foundation of Guangdong Province (Grant No. 2025A1515012950) and the Hebei Major Science and Technology Support Plan (Environmental Management Special Grant, No. 252S3701D).
\item \textbf{Competing interests.} The authors declare no competing interests.
\item \textbf{Data availability.} AtmosCoder-Bench is publicly available at
\url{https://github.com/acodercat/AtmosCoder-Bench}. The repository holds the released corpus in
full: the 436 core problems, each with its verbatim statement, source, executable reference solver,
stored answer and annotations; the 1{,}730 numeric and 2{,}180 paraphrase variants; the 67-item trap
diagnostic together with the shortcut solver and predicted output vector stored for each trap; the
169 scaffolding-ablation pairs; the 131 cross-domain problems; and the imported AtmosSci-Bench MCQ10
set. It also holds the construction, certification and audit artefacts, the per-experiment analysis
documents, the verbatim response traces behind every case study, and the numerical table behind
every figure, so that every value reported here can be traced to the artefact it came from. The complete per-run model outputs, from which all reported
accuracies are recomputed offline, amount to approximately 5\,GB and are publicly available through links provided in the same
repository; they will additionally be deposited in a DOI-issuing public repository on
publication. No material transfer agreements apply to any of these materials.
\item \textbf{Code availability.} All code used to build, certify and evaluate the benchmark is
publicly available at the same repository, \url{https://github.com/acodercat/AtmosCoder-Bench}: the
construction and certification pipeline, including the deterministic faithfulness auditor; the
evaluation harness with its isolated execution sandbox and unit-aware grader; the offline analysis
modules that regenerate every statistic reported here without further model calls; and the scripts
that extract the figure data and render the figures. The code targets Python 3.13 and exact
dependency versions are pinned in the repository's lock file.
\item \textbf{Author contributions.} M.R. and C.M. contributed equally to this work. M.R. and C.M. jointly conceived the project and designed the overall benchmark. M.R. led benchmark construction, including data collection, cleaning, and annotation, and implemented the data and evaluation pipelines. C.M. led the design of the domain-informed evaluation framework, conducted the in-depth analysis and interpretation of the experimental results, and led the writing of the manuscript. Y.Z. provided technical support for experimental implementation and helped with writing the Methods section. D.J. contributed to the preparation and organization of figures, tables, and supporting materials. Y.H. helped with data collection, cleaning, and annotation. M.G. provided supervision and scientific guidance. J.S. provided overall guidance, supervised the project, acquired funding, and critically revised the manuscript. All authors reviewed and approved the final manuscript.
\item \textbf{Correspondence.} Correspondence and requests for materials should be addressed to Jun Song (junsong@hkbu.edu.hk).

\end{itemize}

\bibliography{references}

\clearpage
\includepdf[pages=-]{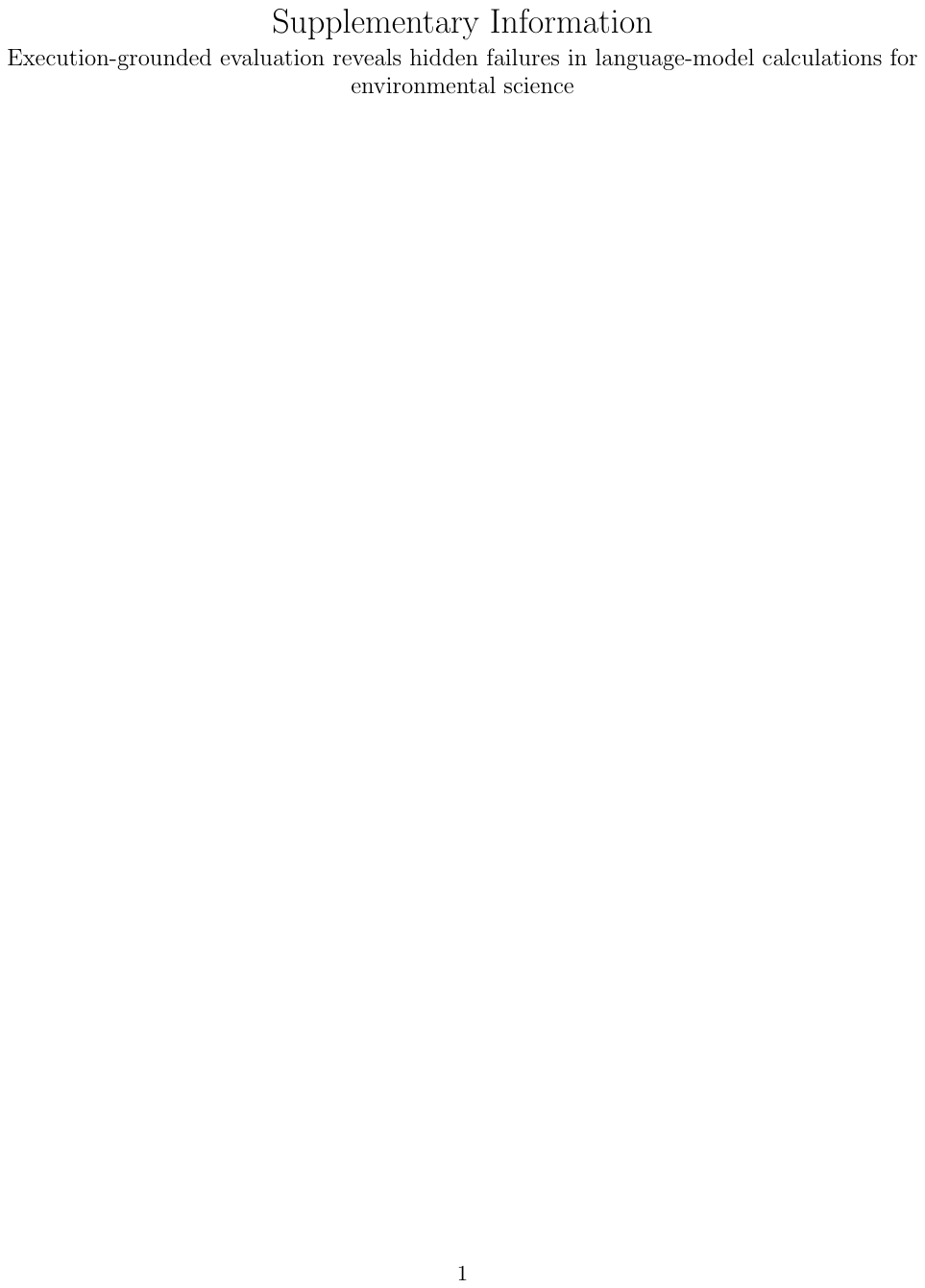}

\end{document}